\documentclass[runningheads]{llncs}
\usepackage[T1]{fontenc}
\usepackage{array}
\usepackage{amsmath}
\usepackage{bbding}
\usepackage{graphicx}
\usepackage{subcaption}
\usepackage{calc} 

\begin{document}
\title{{A Sparsity Predicting Approach for Large}\\ {Language Models via Activation Pattern Clustering}}
%
%
\author{
Nobel Dhar\orcidID{0009-0000-5467-7082} \and
Bobin Deng\orcidID{0000-0001-8361-9025} \and
Md Romyull Islam\orcidID{0000-0003-0550-2248} \and
Xinyue Zhang\orcidID{0000-0002-4243-083X} \and
Kazi Fahim Ahmad Nasif\orcidID{0009-0002-3601-0617} \and
Kun Suo\orcidID{0000-0001-8562-0492}\textsuperscript{(\Envelope)}
}

\authorrunning{N. Dhar et al.}

\institute{
Kennesaw State University, Kennesaw GA 30144, USA \\
\email{
ndhar@students.kennesaw.edu,
bdeng2@kennesaw.edu,
mislam22@students.kennesaw.edu,
xzhang48@kennesaw.edu,
knasif@students.kennesaw.edu,
ksuo@kennesaw.edu}
}

\titlerunning{A Sparsity Predicting Approach for Large Language Models...}   
\maketitle              
\begin{abstract}
Large Language Models (LLMs) exhibit significant activation sparsity, where only a subset of neurons are active for a given input. Although this sparsity presents opportunities to reduce computational cost, efficiently utilizing it requires predicting activation patterns in a scalable manner. However, direct prediction at the neuron level is computationally expensive due to the vast number of neurons in modern LLMs. To enable efficient prediction and utilization of activation sparsity, we propose a clustering-based activation pattern compression framework. Instead of treating each neuron independently, we group similar activation patterns into a small set of representative clusters. Our method achieves up to 79.34\% clustering precision, outperforming standard binary clustering approaches while maintaining minimal degradation in perplexity (PPL) scores. With a sufficiently large number of clusters, our approach attains a PPL score as low as 12.49, demonstrating its effectiveness in preserving model quality while reducing computational overhead. 
By predicting cluster assignments rather than individual neuron states, future models can efficiently infer activation patterns from pre-computed centroids. We detail the clustering algorithm, analyze its effectiveness in capturing meaningful activation structures, and demonstrate its potential to improve sparse computation efficiency. This clustering-based formulation serves as a foundation for future work on activation pattern prediction, paving the way for efficient inference in large-scale language models.

\end{abstract}
\vspace{-9pt}
\section{Introduction}

The fast-paced evolution in transformer-based AI models, such as Large Language Models (LLMs), Large Vision Models (LVMs), and Large Multimodal Models (LMMs), has quickly expanded the smart capabilities to various real-life applications. These large models typically contain several hundred billion or even trillions of parameters—for instance, GPT-4.5~\cite{GPT4.5Wiki} has approximately 12.8 trillion parameters—and must operate on high-performance computing (HPC) systems or large data centers. Centralized processing systems come with significant limitations. First, AI processing requests depend entirely on a stable network connection, requiring end users to transmit their requests to remote HPC systems and wait for the results. Second, as the number of active users grows and AI models become more complex, centralized data centers must adopt increasingly sophisticated designs to maintain the same level of Quality of Service (QoS). However, this will require us to upgrade the complicated cooling system further, which consumes most of the supply power, even more than the computing task itself. Third, remote server processing introduces higher latency and raises concerns about data privacy due to potential security risks during transmission. Therefore, we are motivated to keep a portion of AI tasks on edge devices instead of submitting them to centralized data centers. This edge AI processing on multiple smaller devices essentially increases the AI parallelism for our real-world applications.

\vspace{-0.9pt}
To compress the large AI model and run locally on the resource-constraint edge devices, the leading solutions are quantization \cite{quantization2}, pruning \cite{pruning}, and weight sparsification \cite{sparsification}. Another promising approach is the Mixture of Experts (MoE), which has been adopted in many state-of-the-art LLMs \cite{deepseek}. MoE models improve efficiency by activating only a subset of expert networks at each inference step rather than using all model parameters simultaneously. MoE consists of multiple independent feedforward networks that serve as experts, with a gating mechanism dynamically selecting which ones to use. A small neural network, known as a router, dynamically selects the most suitable experts for each input instance \cite{router2}. This dynamic selection can lower computational costs by only activating the parameters of the most relevant expert. Besides MoE, directly exploring the activation sparsity of pre-trained LLMs can also improve resource efficiency. Unlike MoE, which is explicitly trained with internal expert selection mechanisms, the activation sparsity approach follows the conventional transformer training paradigm. Recent research ~\cite{ndhar2} demonstrates that enforcing activation output with smaller absolute values to zero can obtain additional activation sparsity with negligible accuracy degradation, even in non-ReLU-based models.

By effectively utilizing sparsity, it is possible to activate only a subset of neurons during inference, significantly reducing resource consumption without modifying the model’s original architecture. While MoE models dynamically activate different sets of neurons through a routing mechanism, leveraging activation sparsity in pre-trained models presents a different challenge: identifying which neurons should be active without an explicit selection process. Since these models were not trained with built-in expert selection, determining neuron activations becomes nontrivial. Common approaches attempt to address this by utilizing the similarity of activated channels among semantically related tokens \cite{predictions1}, analyzing activations after the gate projection \cite{prediction2}, or employing a learnable predictor \cite{pmlr-v202-liu23am}. The overall effectiveness of these strategies largely depends on the accuracy of predicting active neurons. However, existing methods have not fully explored the possibility of predicting neuron activations solely based on inputs before computation. This omission is largely due to the immense computational and memory costs associated with directly predicting activations across billions of neurons. Without an efficient mechanism to address this scalability issue, the overhead of such predictions could outweigh the benefits, limiting the practicality of activation sparsity in large-scale models. A more scalable alternative is to cluster activation patterns into a small set of representative centroids, allowing for efficient prediction at pattern level rather than at the individual neuron level. 

In this work, we explore the activation sparsity in Mistral-7B and propose a clustering-based approach to optimize inference efficiency. Instead of predicting activations at an individual neuron level, we identify recurring activation patterns and use centroids to approximate neuron states, significantly reducing the computational burden. To introduce sparsity, we apply a thresholding mechanism proposed by Dhar et al.~\cite{ndhar2}, which enforces 50\% sparsity in the FFN layers of Mistral-7B while maintaining a PPL score of 6.45. This ensures that half of the neurons remain inactive during inference We use the WikiText-2 \cite{wikitext} dataset as the input to the Mistral-7B model to extract activation patterns in the MLP layers for our clustering approach. WikiText-2 is a widely used benchmark dataset that provides a diverse yet relatively clean textual corpus, making it an ideal choice for studying activation behaviors in LLMs.

\begin{figure}[t]
\centering
\includegraphics[width=0.9\columnwidth]{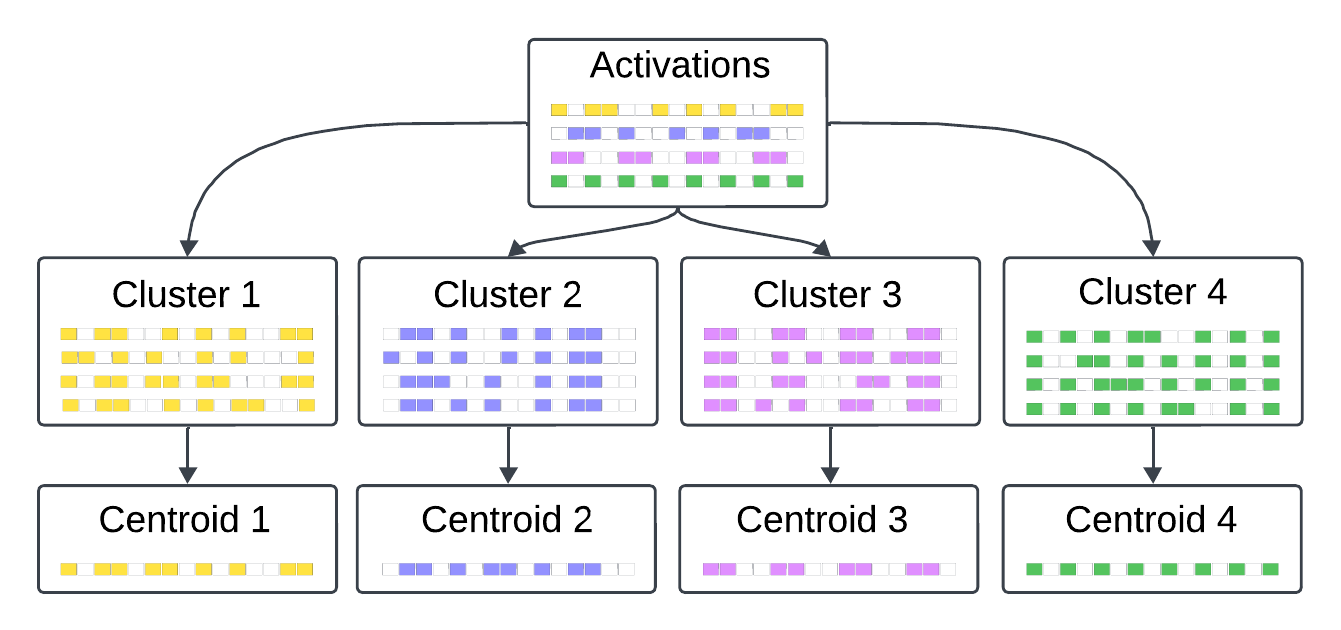}
\caption{Clustering Patterns Into Representative Centroids}
\vspace{-9pt}
\label{fig:clustering}
\end{figure}

Figure~\ref{fig:clustering} provides an overview of our clustering method, illustrating how activation patterns that exhibit similarity are grouped together, forming clusters that share common activation characteristics. Each cluster is represented by a centroid, which is calculated to best approximate the activation status of individual neurons within the group. Each centroid captures a unique activation pattern observed across multiple inputs, allowing us to map future activations to their closest centroid rather than processing them independently. In other words, the centroid serves as a compact representation of the most frequently occurring activation states across multiple patterns. This structured clustering serves as the foundation for our key contributions, which are summarized as follows:
\begin{itemize}
\item We developed a customized clustering algorithm designed specifically to address activation sparsity in LLMs.

\item The proposed algorithm achieves high clustering precision across 10.64 million activation patterns while maintaining minimal perplexity (PPL) degradation, ensuring efficient neuron representation without compromising model performance.

\item We proposed a scalable approach to optimizing sparse activation patterns by clustering billions of neurons into representative centroids, fundamentally reducing the overhead of activation prediction.
\end{itemize}

This work presents a scalable clustering approach to leverage the activation sparsity in LLMs. By grouping similar activation patterns and representing them with centroids, we reduce the computational overhead of activation prediction while preserving the model's accuracy. This method enables efficient inference, optimizing resource usage without altering the model architecture, and serves as a foundational step toward sparse activation-based optimization for LLMs.

The remaining of this paper are structured as follows. Section 2 states our motivations and rationale for selecting binary clustering approaches. Section 3 provides an overview of relevant previous research. Section 4 analyzes the performance of standard clustering algorithms and their limitations in activation sparsity. Section 5 presents the development of our customized clustering algorithm, AWC’, designed specifically for activation sparsity. Section 6 evaluates effectiveness of AWC in terms of clustering accuracy and computational efficiency, and Section 7 concludes the paper with key findings and future directions.

\section{Background and Motivation}
\subsection{The Need for Predicting Neuron Activation Status}

LLMs exhibit activation sparsity, where only a small fraction of neurons activate for a given input, making full computation unnecessary. Despite this, standard LLMs still compute all neurons at every layer, leading to inefficiencies. Predicting active neurons before computation could significantly reduce overhead, but direct prediction at the neuron level is impractical due to the massive number of neurons. A scalable approach to activation prediction is essential to leverage sparsity efficiently for real-world deployment.

\vspace{-9pt}
\subsection{Cost-Effectiveness of Clustering-based Activation Prediction: A Case Study}

LLMs exhibit inherent activation sparsity, where only a subset of neurons is active for a given input. However, predicting individual neuron activations across billions of parameters imposes a substantial computational burden. Given an LLM with \( N \) total neurons, \( L \) layers, \( T \) tokens per sequence, and a per-neuron computation cost of \( C \), the total cost of direct neuron-level prediction is:

\begin{equation}
    \mathcal{C}_{\text{direct}} = N \times L \times T \times C
\end{equation}

For a 7 billion parameter model, approximately \( \frac{2}{3} \) of the parameters belong to the feed-forward network (FFN) layers:

\begin{equation}
    N_{\text{FFN}} \approx \frac{2}{3} \times 7 \times 10^9 = 4.67 \times 10^9
\end{equation}

Since FFN layers consist of three separate projection sub-layers—Gate, Up, and Down—predicting each neuron independently is computationally prohibitive. Instead, our Activation-Aware Patterns Clustering (APC) approach groups activation patterns into a small number of representative clusters, significantly reducing prediction complexity. Assuming \( 2048 \) clusters per sub-layer, the total number of clusters is:
\begin{equation}
    K = 2048 \times 3 = 6144
\end{equation}
This clustering reduces the activation prediction cost to:
\begin{equation}
    \mathcal{C}_{\text{clustered}} = K \times L \times T \times C
\end{equation}
The efficiency gain is:
\begin{equation}
    \frac{\mathcal{C}_{\text{direct}}}{\mathcal{C}_{\text{clustered}}} = \frac{N_{\text{FFN}}}{K} = \frac{4.67 \times 10^9}{6144} \approx 7.6 \times 10^5
\end{equation}
This means our clustering-based activation prediction method reduces computational overhead by a factor of 760{,}000, making activation sparsity prediction computationally feasible. Unlike direct neuron-level prediction, which incurs prohibitive costs, APC enables structured sparsity exploitation while preserving model accuracy, thereby unlocking significant efficiency gains for practical LLM deployments. The clustering algorithm is executed as a pre-processing step and is not involved during inference. Its computational complexity is primarily influenced by the number of activation vectors, the number of centroids, and the centroid update iterations. For our dataset of approximately 10.6 million activation patterns, clustering with 8192 centroids required 36 hours using 8 NVIDIA A100 GPUs. Importantly, this is a one-time offline cost and does not affect runtime inference performance. During inference, no clustering operation is executed. Instead, for each layer of the LLM, one of the precomputed centroids can be predicted based on the input to the LLM. As a result, the cost of clustering itself is completely excluded from the inference runtime. This design significantly reduces the computational load within the MLP layers when compared to dense FFN execution.

\section{Clustering With Standard Binary Clustering Algorithms}
\subsection{Binary Matrix Factorization (BMF)}

In this section, we explore a clustering algorithm called Binary Matrix Factorization (BMF) \cite{bmf} and its performance on the activation values for three projection types—\textit{gate\_proj}, \textit{up\_proj}, and \textit{down\_proj}—across 32 layers. To effectively evaluate the quality of clustering in activation-aware scenarios, we introduce a key metric: clustering precision.

\textbf{Clustering precision} measures how effectively centroids represent the original data points assigned to them. In the context of activation-aware clustering, we focus on the representation of \textit{active neurons} since these are the most significant contributors to computation and the model's performance. Instead of assessing the representation quality over all neurons, our approach evaluates how well the centroids preserve the activation patterns of nonzero neurons.

To quantify clustering precision, we define \(\mathbf{A} \in \{0,1\}^{N \times D}\) as the binary matrix representing the activation states of neurons in the dataset, where \(N\) is the number of data points and \(D\) is the total number of neurons in the model. \( A_{ij} = 1 \) indicates an active neuron at position \( (i,j) \). Similarly, \(\mathbf{C} \in \{0,1\}^{N \times D}\) represents the binary state of the assigned centroids.

We compute the clustering precision as the fraction of correctly represented active neurons:

\begin{equation}
\text{Precision} = \frac{\sum_{i=1}^{N} \sum_{j=1}^{D} (A_{ij} \cdot C_{ij})}{\sum_{i=1}^{N} \sum_{j=1}^{D} A_{ij}}
\label{eq:clustering_precision}
\end{equation}

where the numerator counts the correctly represented active neurons, and the denominator gives the total number of active neurons in the dataset.

This evaluation ensures that clustering quality is assessed based on the most critical neurons rather than all neurons. By focusing on active neurons, we maintain computational efficiency while preserving the accuracy of activation-aware clustering.

\vspace{-10pt}
\begin{table}[ht]
\small
\centering
\renewcommand{\arraystretch}{1.2}
\caption{Clustering Results for BMF}
\label{tab:accuracy_same_clusters_all_layers}
\begin{tabular}{
  |>{\centering\arraybackslash}m{2cm}|
  >{\centering\arraybackslash}m{2.5cm}|
  >{\centering\arraybackslash}m{2.5cm}|
  >{\centering\arraybackslash}m{2.5cm}|}
\hline
\textbf{Projection Type} & \textbf{Total Elements} & \textbf{Clustering Error} & \textbf{Centroid Precision (\%)} \\
\hline
Gate\_proj & 939,524,096 & 275,938,227 & 70.62\% \\
\hline
Up\_proj   & 939,524,096 & 368,293,446 & 60.79\% \\
\hline
Down\_proj & 268,435,456 & 137,573,172  & 48.74\% \\
\hline
\end{tabular}

\end{table}

The clustering was performed using the same set of clusters for all 32 layers. For \textit{gate\_proj} and \textit{up\_proj}, the tensors had a shape of $[2048, 14336]$, and for \textit{down\_proj}, the tensors had a shape of $[2048, 4096]$. The number of clusters was set to \(k = 2048\). Since the features are different among the sub-layers, we used different sets of clusters for the sublayers. The total number of elements processed was 939,524,096 for \textit{gate\_proj} and \textit{up\_proj}, and 268,435,456 for \textit{down\_proj}.

As shown in Table~\ref{tab:accuracy_same_clusters_all_layers}, \textit{gate\_proj} achieved the highest centroid precision of 70.62\% with the lowest error. In contrast, \textit{down\_proj} had the lowest centroid precision at 48.74\% and the highest error percentage. The \textit{up\_proj} demonstrated moderate performance with a centroid precision of 60.79\%. These results suggest that the clustering approach is more effective for the \textit{gate\_proj} layers.

\subsection{Binary-to-Real-and-Back K-Means (BRB-KMeans)}

Binary-to-Real-and-Back K-Means (BRB-KMeans) \cite{BRB-KMeans} is an innovative clustering algorithm tailored for binary data applications in Binary Product Quantization (BPQ). Traditional methods, such as the k-majority algorithm, rely on Hamming distance and majority voting to cluster binary data. However, these approaches often lead to clustering quality degradation, impacting BPQ performance significantly. BRB-KMeans overcomes these challenges by leveraging the high-quality clustering capabilities of k-means in the real-valued vector space. The process begins by transforming binary data into real-valued vectors, applying k-means clustering, and then converting the centroids back into binary format. This transformation capitalizes on the advantages of Euclidean distance and mean-based centroid updates inherent to k-means, providing a significant improvement in clustering quality and BPQ centroid precision.

\vspace{-15pt}
\begin{table}[ht]
\small
\centering
\renewcommand{\arraystretch}{1.2}
\caption{Clustering Precisions for Same Clusters Across 32 Layers}
\label{tab:precision_same_clusters_all_layers_brbk}
\begin{tabular}{
  |>{\centering\arraybackslash}m{2cm}|
  >{\centering\arraybackslash}m{2cm}|
  >{\centering\arraybackslash}m{2cm}|
  >{\centering\arraybackslash}m{2cm}|
  >{\centering\arraybackslash}m{2cm}|}
\hline
\textbf{Projection Type} & \textbf{k = 2048} & \textbf{k = 4096} & \textbf{k = 8192} & \textbf{k = 16384} \\
\hline
Gate\_proj & 69.57\% & 71.90\% & 74.43\% & 77.68\% \\
\hline
Up\_proj   & 59.90\% & 62.57\% & 65.94\% & 70.19\% \\
\hline
Down\_proj & 46.74\% & 50.12\%  & 54.48\% & 60.86\% \\
\hline
\end{tabular}
\end{table}
\vspace{-9pt}
Table~\ref{tab:precision_same_clusters_all_layers_brbk} further explores clustering precision across different layers and projection types—Gate\_proj, Up\_proj, and Down\_proj—for varying cluster sizes (k). The percentage values in this table represent clustering precision for the respective projection types. Gate\_proj consistently demonstrates the highest clustering precision, reflecting its ability to effectively capture the structure of the data. For instance, at k = 16384, Gate\_proj achieves clustering precision of 77.68\%, while Up\_proj and Down\_proj show lower clustering precision values of 70.19\% and 60.86\%, respectively. These results highlight variations in data distribution across different projection types. Gate\_proj stands out for its higher fidelity in representing data points, even with fewer clusters in some cases. The table also reveals that as increases, clustering precision improves across all projection types, demonstrating the algorithm’s adaptability to the complexities of high-dimensional data.

\section{Enhancing Clustering Precision: A Customized Approach}

The previous section discussed our initial approach, where we experimented with two standard binary clustering algorithms to group active and inactive neurons. However, both methods performed similarly and failed to produce acceptable results, as they did not effectively capture the structure of activation sparsity in large language models. The primary limitation was their inability to account for the unique distribution and correlation of active neurons, leading to suboptimal centroid assignments. To address these challenges, we customized the clustering process to better fit the sparsity patterns of activation functions. This led to the development of Activation-Aware Clustering (AWC), a tailored clustering approach designed for efficient and accurate activation sparsity modeling.

\subsection{Assignment}

\begin{figure}[ht]
\vspace{-9pt}
\centering
\includegraphics[width=\columnwidth,trim={0cm 0cm 0cm 0cm},clip]{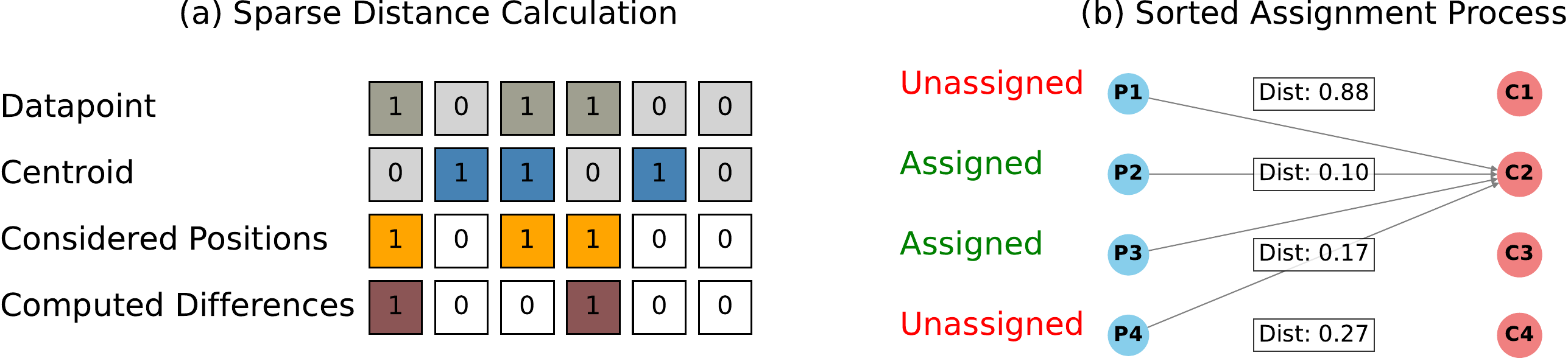}
\caption{Distance Calculation and Assignment Process}
\label{fig:distance}
\end{figure}
\vspace{-9pt}
Figure~\ref{fig:distance}  demonstrates the motivation and methodology behind focusing solely on active neurons (1s) during the assignment phase of clustering. In LLMs, activation patterns are often sparse, with a significant number of inactive neurons (0s) in the high-dimensional space. Traditional distance calculations that treat all positions equally dilute the impact of active neurons, leading to less accurate and less meaningful clustering outcomes.

To address this, we developed a sparsity-aware approach that considers only the active neurons when computing distances between datapoints and centroids. This targeted strategy ensures that the assignment process focuses on the most meaningful dimensions, where the activations contribute to the semantics of the data. By excluding inactive neurons, the computational cost of the assignment step is reduced, allowing the method to scale effectively with the large number of dimensions typically present in LLM activations. Additionally, focusing on 1s aligns the clustering process with the underlying data distribution, as active neurons represent features that are critical for the model’s output. This ensures that centroids are updated to better capture the essential patterns of their assigned datapoints, ultimately leading to more accurate and interpretable clustering results. The sparsity-aware assignment, as shown in the figure, is a fundamental step in addressing the unique challenges posed by LLM activations, improving both efficiency and effectiveness.


Figure~\ref{fig:distance}(b) highlights the mechanism for evenly distributing activation patterns across centroids during the assignment phase. A significant challenge in clustering LLM activations is ensuring that no single centroid becomes overloaded with datapoints, which would result in suboptimal representation of the overall data distribution. Without balancing, a centroid might need to represent a disproportionately large subset of activation patterns, degrading the quality of the clustering.

To address this, the approach sorts unassigned datapoints by their distances to centroids, ensuring that the closest datapoints are assigned first. For each centroid, a fixed number of datapoints is selected based on proximity. In the example shown, centroid C2 receives assignments from datapoints P2 and P3, which are closest to it with distances of 0.31 and 0.43, respectively. This process guarantees that the assignments are not only driven by distances but are also constrained to maintain balanced cluster sizes.

\subsection{Centroid Update}

\begin{figure}[ht]
\vspace{-15pt}
\centering
\includegraphics[width=0.9\columnwidth,trim={0cm 0cm 0cm 0cm},clip]{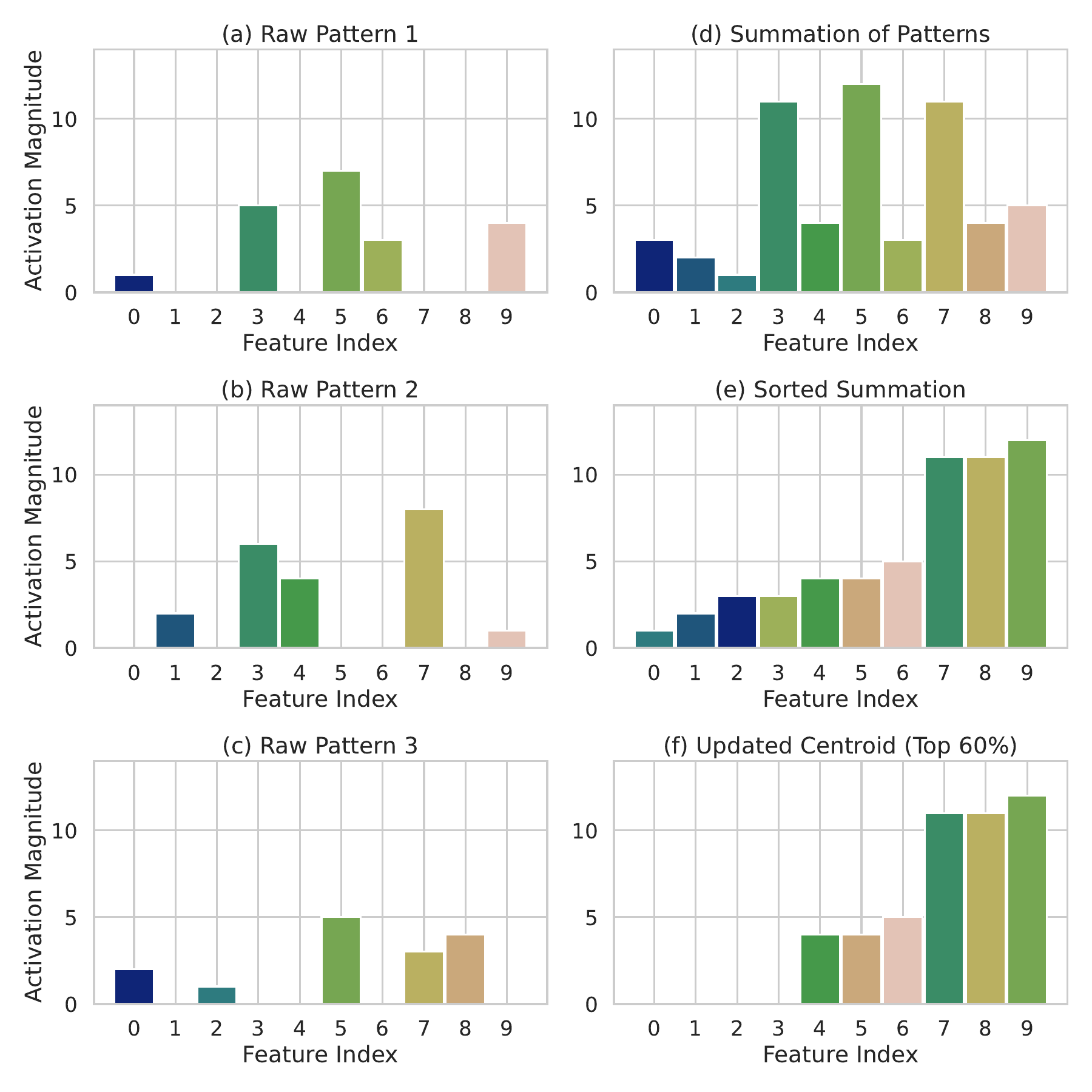}
\caption{Evenly Distributed Assignments}
\label{fig:updating}
\end{figure}
\vspace{-9pt}
The methodology for updating centroids during clustering, as illustrated in Figure~\ref{fig:updating}, is designed to retain the most significant activations (neurons) while achieving a desired level of sparsity. Each cluster’s centroid is calculated by aggregating activation patterns from data points assigned to that cluster. Initially, the raw patterns illustrated in figures 4(a), 4(b), and 4(c) represent the activations of individual neurons for different input tokens. These patterns always include both zero and non-zero values, capturing the sparsity and intensity of the activations.

To compute the centroid, the activation values are summed feature-wise across all patterns in the cluster, as shown in Figure~\ref{fig:updating}(d). This summation aggregates the contributions of individual patterns, creating a single representation that reflects the collective behavior of the activations. Figure~\ref{fig:updating}(e) depicts the next step, where the summed activations are then sorted to identify the most important features. By applying a percentile threshold, the top 60\% of features with the highest activation values are selected to form the updated centroid, as demonstrated in Figure~\ref{fig:updating}(f). This ensures that only the most significant activations are retained while the remaining low-importance features are excluded.

This approach is particularly effective in leveraging the inherent sparsity of LLMs, where only a subset of neurons contributes to any given task. By focusing on the most important activations, the method captures the key information while reducing noise from less relevant features. The percentile-based threshold introduces a consistent level of sparsity across all centroids, ensuring efficiency in both memory usage and computational cost. Moreover, retaining the original activation values for the selected features, rather than binarizing them, helps preserve the intensity information necessary for accurate representation.


\section{Experimental Results}

\begin{figure}[ht]
\vspace{-20pt}
    \centering
    \makebox[\textwidth]{\includegraphics[width=0.9\linewidth, keepaspectratio]{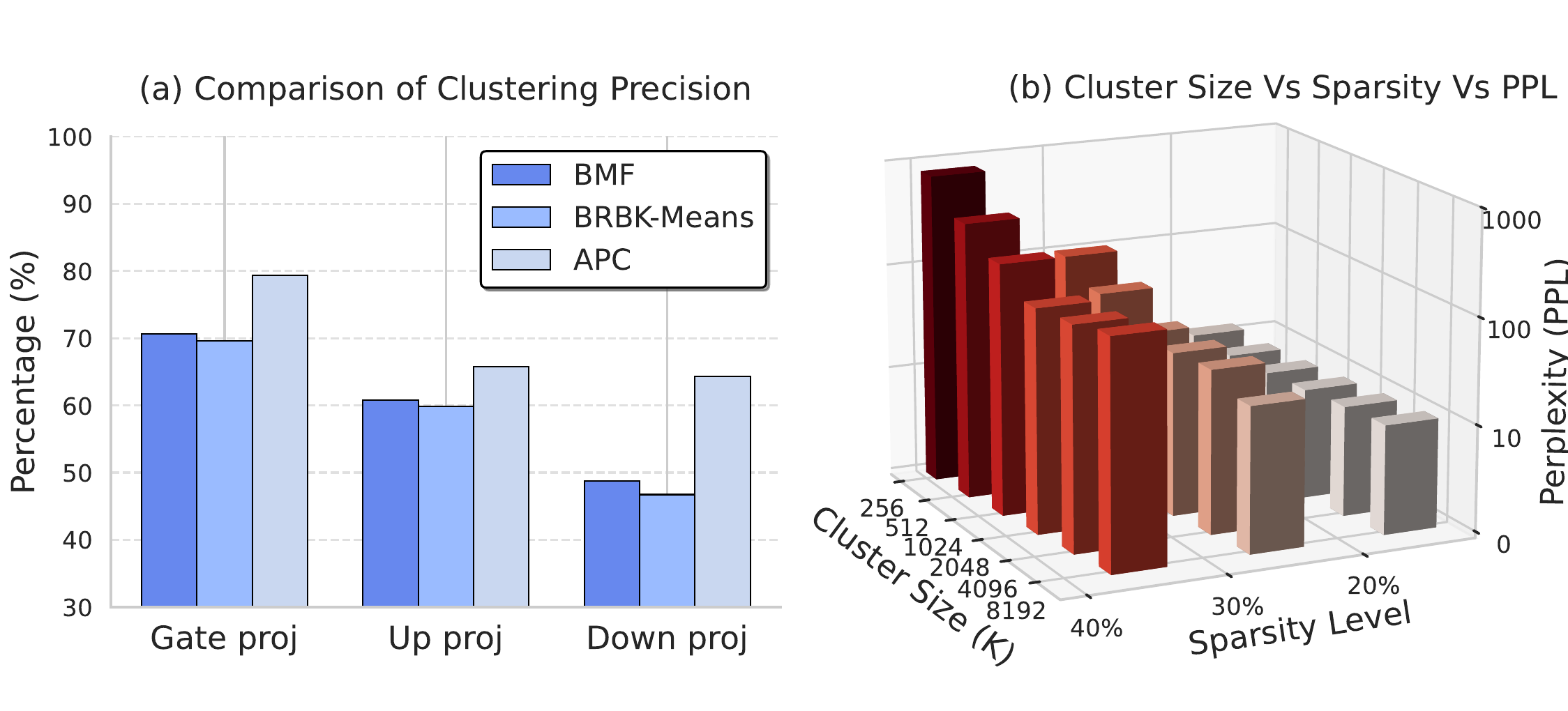}}
    \caption{Results Analysis and Comparison}
    \label{fig:ppl}
\end{figure}

\vspace{-9pt}
\subsection{Evaluation of Clustering Precision}

We applied our customized clustering algorithm to the activation values derived from the input of 163 data points in the WikiText-2 dataset. Each data point consisted of 2048 tokens and went through 32 layers of the LLM. This resulted in a total of $163 \times 32 \times 2048 = 10682368$ activation patterns processed during clustering. The activation values were taken from the \textit{gate\_proj}, \textit{up\_proj}, and \textit{down\_proj} layers separately, with each activation pattern represented by a feature vector of dimensionality 14336. By clustering these high-dimensional activation patterns, the algorithm efficiently compressed and represented the data while maintaining high clustering precision for active features (1's). This experiment validated the effectiveness of our approach in handling large-scale datasets with substantial feature sparsity.


Figure 4(a) illustrates the clustering precision of APC compared to BMF and BRBK-Means across different projection types, highlighting APC’s superior ability to capture activation patterns effectively. The Activation-Aware Patterns Clustering (APC) algorithm demonstrates superior performance compared to traditional clustering methods such as BMF and BRBK-Means, particularly in capturing activation patterns across different projection types. APC achieves the highest centroid precision for the Gate proj (79\%), significantly outperforming BMF (70.62\%) and BRBK-Means (69.57\%). Similarly, for Up proj, APC maintains a notable edge, reaching 65\%, whereas BMF and BRBK-Means lag behind at 60.79\% and 59.90\%, respectively. The most striking improvement is observed in Down proj, where APC achieves 64\%, surpassing both BMF (48.74\%) and BRBK-Means (46.74\%) by a substantial margin. This improvement highlights APC’s ability to better represent neuron activation patterns, making it a more effective clustering strategy for activation sparsity-aware optimization. The results validate APC’s potential to enhance computational efficiency while preserving model accuracy.

\vspace{-9pt}
\subsection{Clustering Precision, Centroid Sparsity and Their Impact on LLM Perplexity}

To evaluate the impact of our clustering-based approach on model performance, we assigned neuron activations based on the centroid of each cluster. We simulated the prediction process with 100\% accuracy in cluster selection, allowing us to assess how using these clusters would affect the model’s PPL score. For reference, the 50\% sparse Mistral-7B model without clustering achieves a PPL score of 6.45 \cite{ndhar2}, which serves as a lower bound when comparing the effectiveness of clustered activation representations.

Figure 4(b) presents the final evaluation of our clustering-based activation sparsity approach, demonstrating its impact on model performance across different cluster sizes (K) and sparsity levels. The experiments were conducted with six cluster sizes (K = 256 to 8192), each tested under three sparsity levels—40\%, 30\%, and 20\%. The key metrics include the number of active neurons in the Gate, Up, and Down projection layers of the feed-forward network (FFN), along with the corresponding PPL scores.  

The key takeaway from this result is that our clustering-based approach maintains low PPL scores even under high sparsity, demonstrating its effectiveness in reducing computational and memory overhead while preserving model performance. Notably, at 40\% sparsity, a moderate increase in cluster size (e.g., from K = 256 to K = 2048) significantly improves PPL, reducing it from 933.78 to 58.84. This trend remains consistent across larger cluster sizes, where higher K compensates for increased sparsity, allowing efficient trade-offs between computational cost and model accuracy. At K = 8192, even at 20\% sparsity, PPL remains as low as 12.49, indicating that our clustering method enables highly sparse yet effective activation representations. 

These results validate the effectiveness of Activation-Aware Clustering (APC) method in preserving model accuracy while reducing computational overhead. By leveraging efficient clustering, our approach ensures that the model retains activation patterns without requiring an excessive number of centroids, thereby striking an optimal balance between performance and computational efficiency. Furthermore, the results suggest higher cluster sizes compensate for increased sparsity, allowing trade-offs between computational cost and model accuracy.

\section{Related Works}
\subsection{Inference Acceleration for LLMs} 

Despite the impressive performance of LLMs, the continuous increase in their size has led to a significant rise in the computational demands for inference, making their deployment challenging \cite{pmlr-v202-liu23am}. To address these high computational costs, various methods for model compression have been developed. These include techniques like quantization \cite{jacob2017quantizationtrainingneuralnetworks}, which reduces the precision of model parameters, pruning \cite{molchanov2017pruningconvolutionalneuralnetworks}, which removes less critical elements, and distillation \cite{tang2019distillingtaskspecificknowledgebert}, which transfers knowledge to smaller models. In addition, efficient sampling methods have been proposed to speed up the decoding process during inference \cite{Wang}. Generally, these acceleration strategies do not utilize the inherent mechanisms within LLMs. However, our approach leverages activation sparsity by clustering similar activation patterns, enabling efficient neuron selection without modifying model weights or altering inference dynamics.

\vspace{-9pt}
\subsection{Leveraging Activation Sparsity for Computational Efficiency}

Recent research has identified both inherent activation sparsity, which naturally emerges in certain LLMs, and forced sparsity, where activations are deliberately constrained using thresholding mechanisms to enhance efficiency \cite{song2023powerinferfastlargelanguage,ndhar2}. Activation sparsity refers to the presence of numerous zero or near-zero values in the activation outputs, which correspond to specific neurons that have minimal impact on the final output for a given input. These neurons can be skipped during the inference process, thereby reducing computational load. Notably, activation sparsity is complementary to other methods like model compression and efficient sampling and can be combined with these techniques to enhance inference speed. DEJAVU \cite{pmlr-v202-liu23am} proposed leveraging activation sparsity in both the Attention and MLP blocks of a Transformer layer using a sparsity predictor composed of two fully connected layers. While this approach is claimed to be less computationally expensive than a nearest-neighbor-based predictor, it still introduces significant overhead. Notably, DEJAVU predicts activation status at the individual neuron level, which scales poorly for large LLMs, as it requires running separate predictors for each MLP layer. Another notable approach, SparseInfer \cite{SparseInfer}, focuses on exploiting activation sparsity in Transformer models but is inherently limited to ReLU-based architectures. The method relies on sign-based comparisons between inputs and weights, which only works for ReLU activations since they naturally produce a large number of zeros. However, modern LLMs predominantly use SiLU or GELU activations, which do not exhibit the same natural sparsity. As a result, SparseInfer requires modifying the activation function of the pre-trained LLM, making it incompatible with non-ReLU-based architectures. In contrast, our proposed Activation-Aware Clustering (AWC) approach addresses these limitations by grouping neurons into activation clusters, allowing activation sparsity to be predicted at the cluster level rather than for each individual neuron. This significantly reduces the prediction overhead while maintaining accuracy. Additionally, AWC does not require any modification to the activation functions of pre-trained LLMs, making it applicable to a broader range of architectures, including those that utilize SiLU and GELU activations.

\section{Conclusion and Future Directions}

In this work, we introduced a clustering-based approach to leverage activation
sparsity in LLMs. By grouping similar activation patterns and predicting cluster
assignments instead of individual neuron states, we significantly reduced the complexity of activation modeling. Our method efficiently captures the underlying activation structures and enables sparse inference without modifying the original model architecture. The experimental results demonstrated that our approach preserves accuracy while achieving substantial reductions in computational cost and memory usage. Specifically, we showed that selecting an optimal sparsity level (20\%) provides the best tradeoff between efficiency and perplexity, making activation sparsity a viable direction for optimizing LLM inference. Additionally, by replacing direct neuron-wise activation prediction with centroid-based cluster assignment, our technique substantially lowers the cost of prediction, making activation sparsity more feasible in large-scale models. 

\vspace{12pt}
\noindent
\textbf{Acknowledgement.}
We are grateful to the anonymous reviewers for their comments and suggestions on this paper. This work was supported in part by U.S. National Science Foundation (NSF) grants SHF-2210744, IIS-2348417 and CNS-2431597. The code is available in the Github repository~\cite{Artifact}.

\vspace{12pt}
\noindent
\textbf{Disclosure of Interests.}
The authors have no competing interests to declare that are relevant to the content of this article.

%
%
%
\bibliographystyle{splncs04}
\bibliography{springer}

\end{document}